\documentclass[10pt,twocolumn]{article}
\usepackage[left=15mm,right=15mm,top=19mm,bottom=25mm]{geometry}
\usepackage{graphicx}
\usepackage{xspace}
\usepackage[font=footnotesize,labelfont=bf]{caption}
\usepackage{hyperref}
\usepackage{url}
\usepackage[capitalise, nameinlink]{cleveref}
\usepackage{cuted}

\newcommand{\verbalinstructions}{\textit{Verbal Instructions}\xspace}
\newcommand{\goalspecifications}{\textit{Goal Specifications}\xspace}
\newcommand{\movementprimitives}{\textit{Movement Primitives}\xspace}

\newcommand{\phin}[1]{\ensuremath{\pi_{#1}}\xspace}
\newcommand{\phinFAST}{\phin{0}{\footnotesize-FAST}\xspace}
\newcommand{\phinKI}{\phin{0.5}{\footnotesize-KI}\xspace}

\title{\textbf{Are Foundation Models the Route to\\\textit{Full-Stack Transfer} in Robotics?}}

\author{
    Freek~Stulp$^{1\ast}$, Samuel~Bustamante$^{1}$, 
    Jo\~{a}o Silv\'erio$^{1}$,\\
    Alin~Albu-Schäffer$^{1}$,
    Jeannette~Bohg$^{2}$,
    Shuran~Song$^{2}$
    \and	
	\small$^{1}$Institute of Robotics and Mechatronics, German Aerospace Center (DLR), Wessling, Germany.\and
	\small$^{2}$Stanford AI Lab, Stanford University, Palo Alto, USA.\\
	\small$^\ast$Corresponding author. Email: Freek.Stulp@dlr.de
}

\date{}

\begin{document}

\maketitle

\begin{strip}
\vspace{-1.5cm}
    
\begin{abstract}\textbf{
In humans and robots alike, transfer learning occurs at different levels of abstraction, from high-level linguistic transfer to low-level transfer of motor skills.
In this article, we provide an overview of the impact that foundation models and transformer networks have had on these different levels, bringing robots closer than ever to ``full-stack transfer''.
Considering LLMs, VLMs and VLAs from a robotic transfer learning perspective allows us to highlight recurring concepts for transfer, beyond specific implementations.
We also consider the challenges of data collection and transfer benchmarks for robotics in the age of foundation models.
Are foundation models \textit{the} route to full-stack transfer in robotics? Our expectation is that  they will certainly stay \textit{on} this route as a key technology.}
\end{abstract}

\vspace*{0.2cm}

\begin{center}
\includegraphics[width=0.8\textwidth]{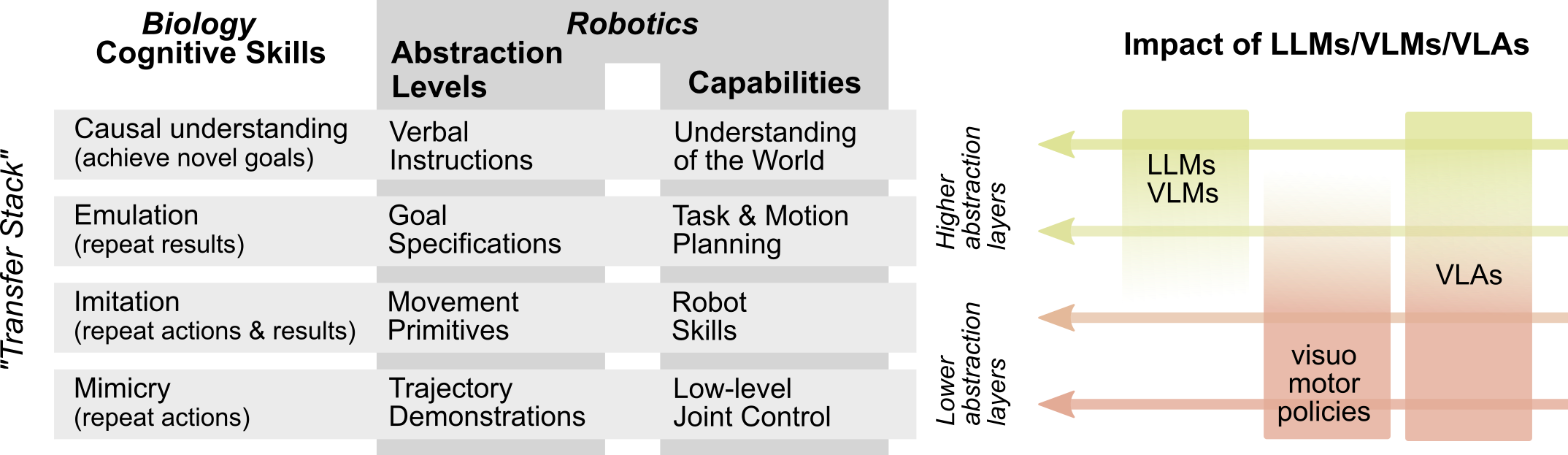}
\end{center}
\refstepcounter{figure}
\footnotesize
\setlength{\parskip}{0pt}
\setlength{\parindent}{0pt}
\textbf{Figure~\thefigure}: Jaquier et al.~\cite{jaquier2024transfer} propose four different levels at which transfer can take place. 
``Full-stack transfer'' means that a robot has the ability to perform transfer at all levels.
As illustrated to the right, several large large AI models, most of them based on transformer networks, have had a profound impact on achieving transfer in different layers of the transfer stack, bringing robots closer than ever to full-stack transfer.
\label{fig:transfer_layers}
\vspace*{0.7cm}
\end{strip}

\section{Introduction}

Humans excel at transferring knowledge between tasks and environments at different levels of abstraction.
We are able to repeat actions (mimicry), repeat results (imitation and emulation), and achieve novel goals based on a fundamental understanding of the world~\cite{jaquier2024transfer}. 
These different levels of abstraction in transfer learning -- proposed by Jaquier et al.~\cite{jaquier2024transfer} -- are illustrated in \Cref{fig:transfer_layers}, where the gray matrix repeats Figure 7 in~\cite{jaquier2024transfer}.  
The center gray column depicts the corresponding level of abstraction at which a task must be specified in robotics; by a human, or generated by the robot itself.  
To successfully perform transfer at a certain task abstraction level, a robot must have all capabilities, from the lowest one up to that level.
Specifying tasks at higher transfer layers is easier for humans, but thus requires more capabilities from the robot. 

Over the last three years, \textit{foundation models} -- especially those based on transformer networks -- have had a profound impact on all layers of the transfer stack.
First, transformer-based large language models (LLMs) and vision language models (VLMs) display a broad and deep linguistic understanding\footnote{Whether LLMs really \textit{understand the world} and have a \textit{causal understanding} (or they are rather like a Chinese Room) is a philosophical question beyond the scope of this paper. To implement the robotic capabilities as higher abstraction levels, it suffices that they mimic this understanding well, e.g. they are able to predict the effects of actions at a language level.} of concepts in our everyday world~\cite{kawaharazuka2025vla}. These foundation models (definition in \Cref{tab:definitions}) acquire this knowledge from the  Internet-scale text and image data that they are trained on. They provide robots with queryable models to realize transfer in the higher layers of the transfer stack.

Second, novel visuomotor policies such as diffusion policies~\cite{chi2023diffusionpolicy} and flow matching~\cite{lipman2023flowmatching,black2024pi0} are able to acquire versatile skills from task demonstrations through imitation learning~\cite{schaal1999isimitationlearning}. Especially transformer networks have provided these models with the ability to represent error-recovery strategies and to transfer across task variations at lower transfer layers~\cite{trilbmteam2025carefulexaminationlargebehavior}.

\begin{figure}
    \centering
    \includegraphics[width=1.0\linewidth]{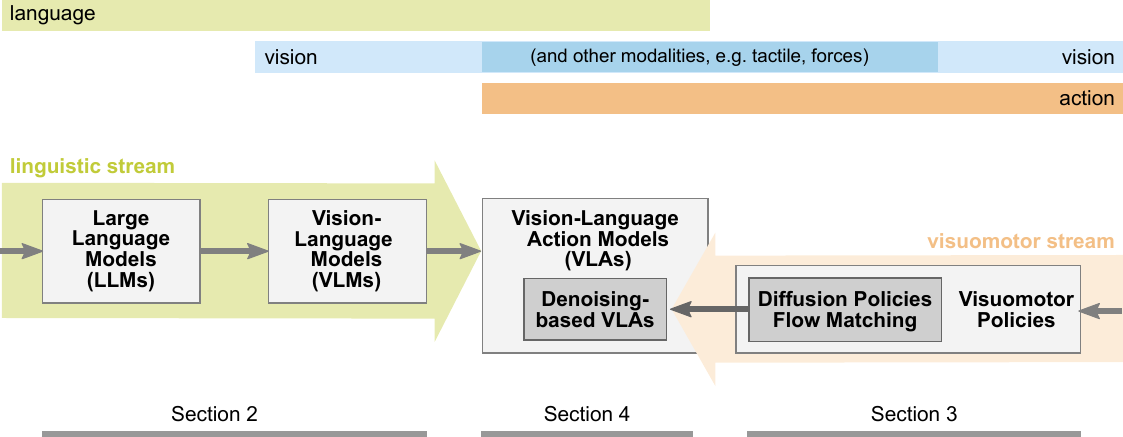}
    \caption{Transformer networks have had a profound impact on the performance of LLMs/VLMs (\Cref{sec:llmvlm}) and the ability to learn  visuomotor policies, e.g.  diffusion policies, from demonstrations (\Cref{sec:visuomotor}).
    VLAs (\Cref{sec:vlas}) arose as extensions of VLMs. 
    Denoising-based VLAs (\Cref{sec:gradient_stopping}) arose by merging diffusion policies or flow matching with VLMs, which was made possible by the shared modality of vision.}
    \label{fig:streams}
\end{figure}

In the robotics community, pre-trained VLMs were extended and fine-tuned with action data to yield Vision Language Action models (VLAs, definition in \Cref{tab:definitions}).
As visuomotor policies and VLMs share the vision modality and are both frequently based on transformer networks, these two research streams could naturally be merged, as illustrated in \Cref{fig:streams}. This led to a particular class of VLAs which we call \textit{denoising-based VLAs}, which correspond to the VLA classes 5/6/7 in Fig. 4 of \cite{kawaharazuka2025vla}.
By using language for transfer at high levels of abstraction and novel policy representations at low levels within one architecture, VLAs have brought robots closer to achieving transfer at all abstraction levels, i.e. ``\textit{full-stack transfer}'', than ever before.

This article is primarily targeted at roboticists who are interested in understanding the impact of foundation models and transformer networks on transfer learning in robotics.
Our main aims are 
1)~to provide an overview of recent works that specifically consider the impact of model design decisions on transfer, e.g. through empirical evaluations.
2)~to map VLA model components to the transfer layers in \Cref{fig:transfer_layers}, to identify what types of transfer take place where in the models.
3)~to highlight limitations and open questions related to VLA architectures, data collection, and benchmarking in the context of transfer learning.  
Works cited in this article have been selected according to these aims.
For readers interested in a recent review on VLAs that aims at completeness, we recommend~\cite{kawaharazuka2025vla}.

The main structure of this article follows \Cref{fig:streams}. The linguistic and visuomotor streams are described in \Cref{sec:llmvlm} and \Cref{sec:visuomotor} respectively, followed by VLAs in \Cref{sec:vlas}. Challenges in data collection and the benchmarking of transfer performance in the age of foundation models are presented in \Cref{sec:data} and \ref{sec:benchmarking}. We conclude with \Cref{sec:conclusion}.

\begin{table}[ht]
  \centering
\begin{small}
  \begin{tabular}{|p{0.95\columnwidth}|}
  \hline
  \textbf{Transfer Learning} \textit{``leverages prior knowledge from a source space, composed by a robot, a task, and an environment, to improve the performance in a target space, where one or more modes differs from the source space.''}~\cite{jaquier2024transfer}\\[0.2cm]
  
  \textbf{Foundation models} are \textit{``trained on broad data (generally using self-supervision at scale) that can be adapted to a wide range of downstream tasks.''}~\cite{bommasani2021foundationmodels}. This definition applies to all LLMs, VLMs and VLAs cited in this paper.\\[0.2cm]

  A \textbf{Vision Language Action model (VLA)} \textit{``is a system that takes visual observations and natural language instructions as required inputs and may incorporate additional sensory modalities. It produces robot actions by directly generating control commands.''}~\cite{kawaharazuka2025vla}. In \Cref{sec:vlas}, we adhere to the classification of VLA architectures in Fig. 4 in~\cite{kawaharazuka2025vla}.\\
  \hline
  \end{tabular}
\end{small}
  \caption{\label{tab:definitions} Definitions used in this article.}
\end{table}

\section{Transfer in Higher Layers: LLMs and VLMs}
\label{sec:llmvlm}

In the context of the transfer layers defined in \cite{jaquier2024transfer}, LLMs and VLMS have had a profound impact in particular on the \verbalinstructions and \goalspecifications layers in \Cref{fig:transfer_layers}. We now provide a very brief LLM-VLM-VLA history, with a focus on these layers.

Shannon observed \textit{``that it is perhaps reasonable to represent English text as a time series produced by an involved stochastic process.''}~\cite{shannon1950redundancy}. Based on this observation, Shannon proposed an algorithm for generating English sentences by iteratively sampling words from a probability distribution conditioned on the last $n$ words in the sentence being produced. 
This autoregressive algorithm is still the backbone of current LLMs, VLMs and VLAs. Scaling the algorithm up to achieving modern-day LLM performance relied on: 1)~data-driven learning of word embeddings in a vector space; 2)~the abundance of training data on the Internet; 3)~the use of transformer networks to represent the probability distribution~\cite{vaswani2025attention}.
It was shown that LLMs are unsupervised multi-task learners that excel
in transfer across tasks, and that they are able to outperform specialist models~\cite{radford2019language}. 

Vision encoders such as CLIP~\cite{radford2021clip} enabled LLMs to process both word embeddings \textit{and} image patch embeddings, leading to multi-modal LLMs called Vision Language Models (VLMs).
As VLAs build on VLMs, and VLMs have been trained on Internet-scale text and image data, VLAs have access to the common-sense knowledge represented in the VLMs.  
Early anecdotal evidence highlighting the task transfer abilities of VLMs in VLAs includes zero-shot success on achieving the task ``move coke can to [a picture of] taylor swift''~\cite{brohan2023rt2}. Open-vocabulary pick-and-drop tasks in real homes~\cite{Liu_2024okrobot} highlights this ability at scale.
Ablation studies have repeatedly corroborated the importance of using web data to pre-train the VLM backbone in VLAs~\cite{black2024pi0,nasiriany2024rtaffordance,black2025pi05}.
Furthermore, it has been shown that fine-tuning VLMs for improved spatial reasoning can improve performance for robot control~\cite{kim2024openvla,qu2025spatialvla,geminiroboticsteam2025}.

An interesting feature of the VLA \phin{0.5} -- to be discussed in more detail in \Cref{sec:gradient_stopping} -- is that it uses a fine-tuned Paligemma VLM to internally generate intermediate language subtasks (e.g. ``pick up the pillow'') from overall top-level language instructions (e.g. ``clean the bedroom'')~\cite{black2025pi05}. 
This intermediate representation corresponds to the \goalspecifications layer between \verbalinstructions and \movementprimitives.
As highlighted in~\cite{black2025pi05}, these different layers can be trained with different datasets, leading to more effective use of training data and thus better transfer. For instance, the \goalspecifications layer can be trained independently of the others by using examples from the web or with data from human ``supervisors'' that guide the robot through a complex tasks step by step~\cite{black2025pi05}. Learning the specific actions to achieve a subtask is left to the \textit{Robot Skill} layer, which is discussed in the next section.

\section{Transfer in Lower Layers: Visuomotor policies}
\label{sec:visuomotor}

Early approaches to perception-driven robot control were built around visual servoing, which relied on handcrafted visual features and geometric models to compute control commands directly from images \cite{chaumette2006visualservoing_basic, chaumette2007visualservoing_advanced}. In parallel, the development of movement primitives \cite{ijspeert2013dmp} provided a way to close the control loop on low-dimensional proprioceptive variables, enabling robots to reproduce learned skills with structured, state-dependent feedback. These approaches established the importance of coupling perception and action, but they required either precise models or carefully engineered state representations, which strongly limited their scalability.

The emergence of visuomotor policy learning in the mid-2010s~\cite{levine2016endtoend,kalashnikov2018qtopt} represented a major shift, as robots began learning end-to-end mappings from images to actions. However, these approaches were either highly data-intensive --- particularly when targeting multi-task learning ~\cite{kalashnikov2018qtopt} --- or relied on task-specific instrumentation to achieve data efficiency in single-task settings~\cite{levine2016endtoend}. Their predominant reliance on reinforcement learning further limited scalability and real-world applicability.

Recent advances in robot learning and transformer networks have, for the first time, enabled the training of visuomotor policies from human demonstrations at realistic time and data scales. %
A key factor enabling this capability is the use of diffusion policies (DPs) \cite{chi2024diffusionpolicy}, which adapt denoising diffusion processes 
to the computation of robot control actions. %
Closely related flow-matching formulations \cite{lipman2023flowmatching} instead learn a continuous-time transport from noise to actions, offering a complementary generative perspective with similar representational power.
Rather than modeling the joint distribution over observations and actions, these approaches learn conditional distributions of actions given observations.

The success of DPs can be attributed to two main factors. On the one hand, diffusion-based denoising helps stabilize learning and improves robustness to out-of-distribution states, when compared to classical, deterministic function approximators. As a consequence, DPs can be trained with significantly smaller datasets when compared to other popular deep-learning-based approaches to behavior cloning and imitation learning, as they are not restricted to explicit parametric families of distributions. %
On the other hand, although diffusion policies were initially proposed with both convolutional neural networks and transformer networks~\cite{chi2023diffusionpolicy}, transformer-based models have emerged as particularly promising in DP-based VLA implementations (see \Cref{sec:vlas}) due to their strong temporal modeling --- facilitating long-horizon tasks --- high sensory expressivity, and reliable scaling with data and compute.

\subsection{Types of Transfer in Visuomotor Policies}  

When trained to minimize losses in Cartesian space (e.g., end-effector poses or velocities), visuomotor policies operate at the level of \textit{Imitation} (\Cref{fig:transfer_layers}).
Accordingly, cross-task and cross-embodiment transfer primarily occurs at the level of \movementprimitives, where learned actions are closely tied to demonstrated outcomes.
Notably, results such as \cite{trilbmteam2025carefulexaminationlargebehavior} suggest that transfer can also occur at the level of \textit{Emulation}, where task outcomes are reproduced in novel situations despite differences in the executed actions.
One possible explanation is that, unlike in \cite{chi2024diffusionpolicy}, the incorporation of language instructions in the observation space in \cite{trilbmteam2025carefulexaminationlargebehavior} narrows the action distribution to modes consistent with the intended goal. %
In contrast, several VLA frameworks learn joint-space commands directly, positioning them at the level of \textit{Mimicry} and inherently limiting their ability to reproduce outcomes across embodiments, see \Cref{sec:cross_embodiment} for a detailed discussion.

A noticeable trend in recent VLA architectures -- to be discussed in more detail in \Cref{sec:vlas} -- is that increasing the diversity of training data directly improves generalization. This is exemplified by the progression from RT-2 to RT-X, where incorporating demonstrations from a broader set of robots, labs, and embodiments led to better transfer to unseen tasks and environments~\cite{collaboration2024openx}. Although these models are not diffusion-based, the same principle appears to carry over to diffusion and flow-based policies, as well as VLAs with diffusion or flow-based action heads.
This trend is analyzed in \cite{trilbmteam2025carefulexaminationlargebehavior}, which shows that fine-tuning a Diffusion Transformer pre-trained on a diverse set of tasks leads to statistically significant improvements in success and task completion rates under both in-distribution and out-of-distribution conditions.

\section{Towards Full-stack Transfer: Vision Language Action Models}
\label{sec:vlas}

The analogy has been made that if LLMs are akin to the prefrontal cortex, and the vision encoders of VLMs to the visual cortex, then VLAs add a motor cortex to the model~\cite{driess2025knowledgeinsulating}.
To do so, most VLA approaches extend a VLM with a dedicated detokenizer, action head, or action expert. 

In this section, we consider two questions that are pertinent to transfer learning in VLAs:
1) How is the action head/expert or detokenizer implemented?
2) Should all layers be trained simultaneously, sequentially, or independently?
Our aim is to highlight works that provide empirical evidence for the impact of different answers to these questions on transfer learning\footnote{Our aim is not to provide an exhaustive overview of current VLA architectures, which  \cite{kawaharazuka2025vla} already provides.}.

This section is structured as follows:
We first introduce VLAs with discrete action tokens as a baseline. This helps to understand more complex VLA architectures, and motivates the need for more sophisticated action detokenizers/action experts/action heads.
In \Cref{sec:action_compression}, we describe the impact of action compression on transfer.
We then consider the impact of \textit{knowledge insulation} with \textit{gradient stopping} in denoising-based VLAs (\Cref{sec:gradient_stopping}) or through \textit{in-painting} (\Cref{sec:in_painting})\footnote{Prioritizing a structure that emphasizes impact on the transfer layers entails VLA models not always being introduced in chronological order.}.
Finally, in \Cref{sec:cross_embodiment}, we consider cross-embodiment as a transversal topic across different VLA architectures.

\subsection{VLAs with Discrete Action Tokens}
\label{sec:vla_dat}

\begin{figure*}
  \centering
  \includegraphics[width=0.9\textwidth]{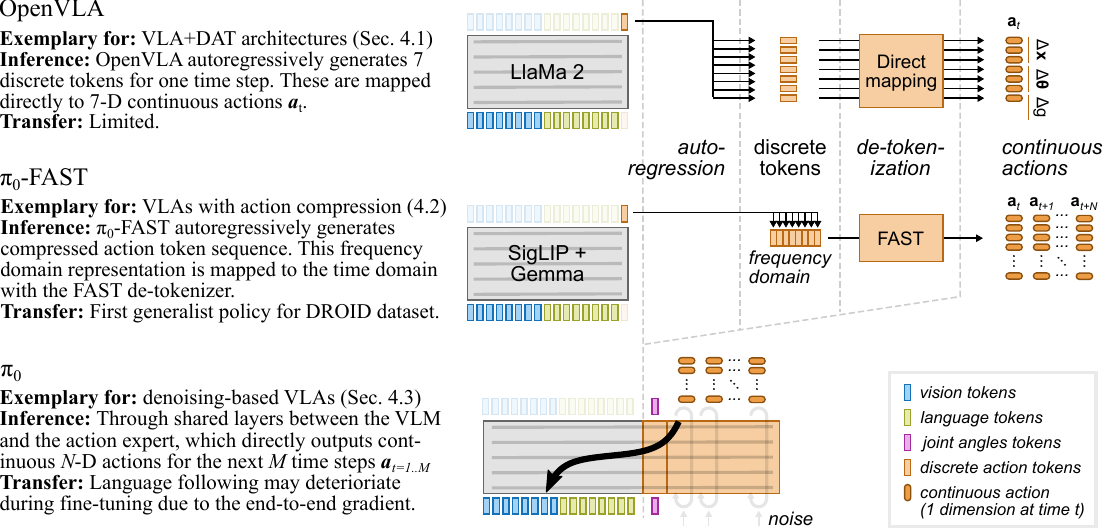}
  \caption{\small \label{fig:action_compression} 
  Exemplary implementations for three distinct classes of VLAs, according to the classification in~\cite{kawaharazuka2025vla}: 
  $\bullet$ OpenVLA~\cite{kim2024openvla}, see \Cref{sec:vla_dat}; 
  $\bullet$ \phinFAST~\cite{pertsch2025fast}, see \Cref{sec:action_compression};
  $\bullet$ \phin{0}~\cite{black2024pi0}, see \Cref{sec:gradient_stopping}. The order follows the structure of the article, and is not chronological.}
\end{figure*}

VLAs with Discrete Action Tokens, denoted ``VLA+DAT'' in~\cite{kawaharazuka2025vla}, essentially \textit{are} VLMs whose discrete output tokens map directly to actions. This class of VLAs includes models such as RT-2~\cite{brohan2023rt2} and OpenVLA~\cite{kim2024openvla}.
In OpenVLA for instance, the values of 7 discrete output tokens of the VLM are directly mapped to a 1-D gripper command and the continuous 3-D translation and 3-D rotational movements of the robot end-effector~\cite{kim2024openvla}, as illustrated in \Cref{fig:action_compression}.

VLA-DAT models are trained by fine-tuning a pre-trained VLM with action data, e.g. human demonstrations containing language commands, vision data, and robotic trajectories, see \Cref{sec:data}.
During the training process, the semantics of output tokens that initially represented English language are re-learned to represent discrete actions.
As~\cite{driess2025knowledgeinsulating} summarize, it's ``\textit{like controlling your arm by verbally saying which muscles should contract}''.

In practice, the  transfer ability of VLA-DATs\footnote{We note that autoregressive VLAs also have several disadvantages such as inefficient inference~\cite{kim2025OFT,driess2025knowledgeinsulating} and the lossiness of the action representations~\cite{driess2025knowledgeinsulating}. Important though they are, we do not explore these aspects further so as to focus on our primary topic of transfer learning.}
is limited~\cite{pertsch2025fast}. 
Moreover, as VLA-DATs arise from models made for text outputs --- and not robot actions---, more advanced detokenizers, action heads and action experts were introduced later as separate modules in VLAs, i.e. effectively insulating the knowledge needed for the \textit{Imitation} and \textit{Mimicry} layers from the \textit{Emulation} and \textit{Causal understanding} layers.
These modules are introduced in the next subsections. 

\subsection{Action Compression with Tokenizers}
\label{sec:action_compression}

Tokenizers such as FAST~\cite{pertsch2025fast} and BEAST~\cite{zhou2025beast} use Fourier transforms and B-splines respectively to compress and tokenize a sequence of continuous actions over time into a sequence of action tokens.
In FAST, byte pair encoding is used as an additional compression step~\cite{pertsch2025fast}. 
These action token sequences are used to fine-tune the VLM. This corresponds to the VLA architecture 4 in Fig. 4 of \cite{kawaharazuka2025vla}.
During inference, autoregression is used to generate token sequences, which a detokenizer converts into a continuous action sequence, see \Cref{fig:action_compression}. 

The impact of action compression was highlighted by the milestone that \phinFAST was the first VLA to be able to learn a generalist policy on the full DROID dataset~\cite{pertsch2025fast} and still follow language commands zero shot in novel environments. 
Furthermore, using a frequency domain representation enables FAST to adapt better to actions that change at different frequencies than the fixed discretization times used in for instance OpenVLA. For this reason, the improvement in transfer learning with FAST is especially pronounced for datasets in which actions change at a high frequency, such as DROID~\cite{pertsch2025fast}. 

\subsection{Knowledge Insulation through Gradient Stopping}
\label{sec:gradient_stopping}

After VLA-DAT architectures and VLAs with action compression, we now consider denoising-based VLAs, which resulted from the merging of the two streams illustrated in \Cref{fig:streams}. Such VLAs combine a VLM with
an denoising-based action expert (e.g. a diffusion policy or flow matching) to implement the lower layers of the transfer stack. Such VLAs correspond to the sensorimotor architectures classes 5/6/7 in \cite{kawaharazuka2025vla}
\phin{0} is an exemplary VLA that is depicted in \Cref{fig:action_compression}.

When implemented as a transformer network,
the action expert can be very tightly integrated with a transformed-based VLM, enabling self-attention~\cite{black2024pi0} or cross-attention~\cite{nvidia2025gr00t} between neural layers in the VLM and the action expert.
The advantages of this approach are:
1) only one VLM forward pass may suffice to generate a discrete action token, avoiding expensive autoregression as needed for detokenizers;
2) despite the tight integration, the action expert can be called at a different frequency from the VLM.
3) the action expert can attend to vision data in the VLM;
4) the action expert and VLM can be fine-tuned simultaneously end-to-end, allowing intermediate discrete actions to represent the information necessary to transfer well across all tasks in the dataset for fine-tuning;

However, the latter point is a pro with a con, as it has been shown that using action data for end-to-end fine-tuning of denoising-based VLAs can degrade the language following abilities of the VLM backbone~\cite{driess2025knowledgeinsulating}. 
In the context of the layers in \Cref{fig:transfer_layers}: fine-tuning on action data to increase performance on lower motor layers has been shown to decrease transfer performance at higher layers.
This may have a severe impact on transfer learning; imagine if practicing your tennis backhand somehow decreased your ability to understand instructions from your trainer!

Several approaches have been proposed to address this issue. First, the weights of the VLM can simply be frozen during fine-tuning on action data~\cite{nvidia2025gr00t}. This avoids the con, but also precludes the pro of the VLM and action experts being tailored to each other~\cite{driess2025knowledgeinsulating}. 

A variety of other solutions have been proposed during the evolution of the \phin{0.*} family of VLAs\footnote{The variety of solutions and their thorough empirical evaluation on transfer make the \phin{0.*} family particularly relevant to this article. As mentioned in the introduction, readers also interested in a more exhaustive review of VLAs  may consider~\cite{kawaharazuka2025vla}.}, which are illustrated in \Cref{fig:action_compression} and \ref{fig:knowledge_insulation}, and which we summarize as follows, now in chronological order:

\begin{figure*}
  \centering
  \includegraphics[width=0.9\textwidth]{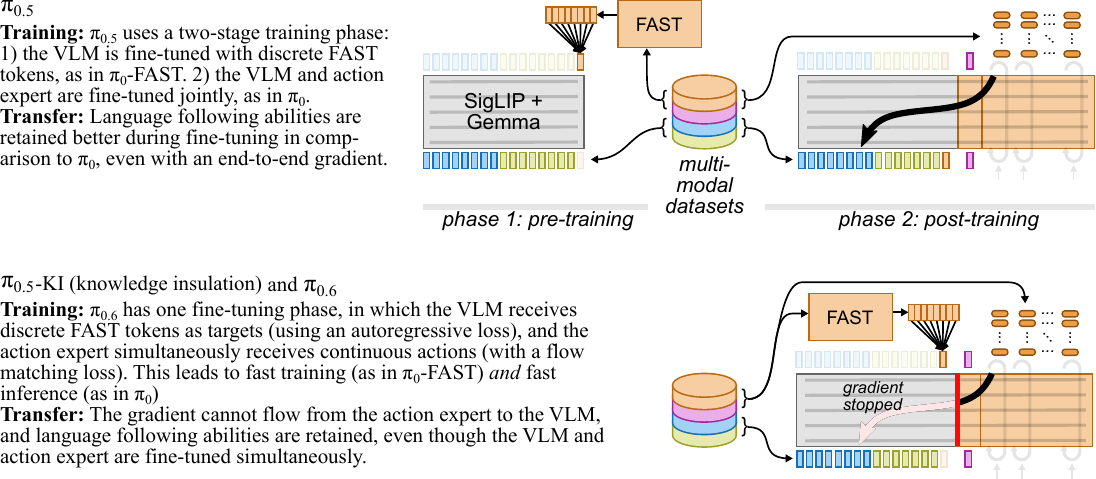}
  \caption{\small \label{fig:knowledge_insulation} Overview knowledge insulation implementations in \phin{0.5}~\cite{black2025pi05}, \phinKI~\cite{driess2025knowledgeinsulating}, and \phin{0.6}~\cite{amin2025pi06vlalearnsexperience}, which are implementations of denoising-based VLAs.
  Whereas \Cref{fig:action_compression} showed \phin{0} and \phinFAST and focussed on inference, this figure rather depicts different training phases.}
\end{figure*}

\begin{description}

\item[End-to-end fine-tuning: Problem.]
\phin{0}~\cite{black2024pi0} consists of a VLM with an action expert based on flow matching, which attends the VLM through self-attention in shared layers. When fine-tuning, propagating the gradient end-to-end through both can deteriorate language following capabilities in higher transfer levels of the VLA~\cite{driess2025knowledgeinsulating}.

\item[Solution 1: Action compression.]
\phinFAST~\cite{pertsch2025fast} is an autoregressive VLA. As mentioned in \Cref{sec:action_compression}, action compression was essential to ensuring generalization across many tasks whilst retaining language-following abilities.

\item[Solution 2: Two training phases.]
\phin{0.5}~\cite{black2025pi05} combines ideas from \phin{0} and \phinFAST, and introduces two training phases illustrated in \Cref{fig:knowledge_insulation}.
In the second stage, the flow matching action expert essentially learns to perform the task of the FAST-detokenizer used in \phinFAST. By decoupling the discrete and continuous training phases, language following abilities are retained. 

\item[Solution 3:] \textbf{Knowledge insulation through gradient stopping.}
\phinKI~\cite{driess2025knowledgeinsulating} and \phin{0.6}~\cite{amin2025pi06vlalearnsexperience} merge the two training phases in \phin{0.5} into one, providing FAST discrete action tokens to the VLM backbone during fine-tuning, whilst simultaneously providing continuous actions to the action expert, see \Cref{fig:knowledge_insulation}. 
However, the gradient flow from the action expert to the pre-trained weights in the model is stopped; a technique called ``knowledge insulation'' (KI). The action expert can still attend the VLM layers during inference, but thus cannot change them during training.
This avoids the deterioration of language following observed in \phin{0}, and leads to faster training and inference than in \phin{0.5} (Solution 2).
\end{description}

Beyond the specific implementation details of the \phin{0.*} family of VLAs, we believe that the main lesson learned from their evolution is that naively connecting and fine-tuning denoising-based action experts may have unexpected negative side effects on transfer. Knowledge insulation between the VLM and the action expert -- the upper and lower layers of the transfer stack in \Cref{fig:transfer_layers} -- must therefore be actively considered, and several approaches have been explored.

\subsection{Knowledge Insulation through In-painting}
\label{sec:in_painting}

In~\cite{driess2025knowledgeinsulating}, ``knowledge insulation'' refers specifically to gradient stopping. In the context of the transfer stack in \Cref{fig:transfer_layers}, we broaden the definition, and consider knowledge insulation to be any approach in which transfer layers are trained independently, without training signals being exchanged between them.
This definition also applies to VLM-based architectures that use \textit{in-painting}~\cite{yuan2024robopoint,huang2024rekep,nasiriany2024rtaffordance,li2025hamster,singh2025ogvla}, In these approaches, a VLM generates textual instructions to ``paint'' additional information on top of the input image, e.g. paths~\cite{li2025hamster}, keypoints~\cite{huang2024rekep}, or grasp poses~\cite{nasiriany2024rtaffordance}. These annotated images are then passed to an action expert\footnote{The VLA definition in \Cref{tab:definitions}  applies to the models in  \Cref{sec:vla_dat}, \ref{sec:action_compression}, and \ref{sec:gradient_stopping}. Whether it applies to the models based on in-painting is debatable. However, for the structure of the article, it is advantageous to include them in \Cref{sec:vlas}.}.

An example is OG-VLA~\cite{singh2025ogvla}, which generates actions in in three phases: 
1)~a fine-tuned VLM generates a token description of the next end-effector via-point and the gripper commands; 
2) this description is painted on top of an orthographic rendering of the input image; 
3) the action expert takes the in-painted image and further inputs, and generates continuous actions. 
This method shows high performance in transfer to novel objects and scenes in an ARNOLD benchmark, outperforming \phinFAST and \phin{0.5}.

The explicit distinction between \textit{what} to do (VLM-based reasoning) and \textit{how} to do it (execution by the action expert)~\cite{shridhar2022cliport},
allows the VLM and action expert to be fine-tuned independently.
And the reasoning abilities of VLMs can be fine-tuned with the need for costly action data, also leading to better overall transfer~\cite{li2025hamster}.
This again highlights the advantages of knowledge insulation between different transfer layers in~\Cref{fig:transfer_layers}.

\subsection{Approaches for Cross-embodiment Transfer in VLAs}
\label{sec:cross_embodiment}

But the \textit{Mimicry} layer refers to repeating actions at the level of (joint) trajectories, which are specific to embodiments, potentially hindering cross-embodiment transfer.
While there has been evidence of \textit{positive transfer}, i.e. \textit{``co-training on data collected on multiple robots improves performance on the training task''}~\cite{collaboration2024openx,geminiroboticsteam2025geminirobotics15pushing}, evidence that VLAs can solve tasks zero-shot on novel robotic embodiments is scarce.

Robot hardware can differ in many respects, posing different challenges for embodiment transfer. Example differences include  
1) the \textit{robot kinematics}, dynamics, and and actuation space,
2) the camera configuration and placement, i.e. \textit{hand-eye transforms},
3) the \textit{visual appearance} of the arm and gripper in the camera images, 
4) \textit{different input modalities} such as tactile sensing.

A common approach to achieving embodiment transfer between robots with different \textit{kinematics} is for the VLA to output actions in end-effector space, and to use an inverse kinematics model of the robot to map these to joint angle space~\cite{kim2024openvla,collaboration2024openx}.
A second approach is to train embodiment-specific action decoders, one for each embodiment~\cite{nvidia2025gr00t}. In these two approaches, the \textit{Mimicry} layer in \ref{fig:transfer_layers} is implemented as a separate module, and transfer is achieved by using a robot-specific model -- physics-based or learned from data -- to implement the \textit{Mimicry} layer.
A third approach is to provide information about the robot kinematics as an input to the VLA, and fine-tune the VLA to adapt the action tokens accordingly. 
For instance, RDT uses a heterogeneous input format which embeds different aspects of the trained kinematic chains~\cite{liu2025rdtb}, and 
X-VLA uses sets of learnable tokens to distinguish between different embodiments and action spaces~\cite{zheng2025xvlasoftpromptedtransformerscalable}.

A pragmatic approach to reducing differences in \textit{hand-eye transforms} between different robots is to attach the eye close to the hand, i.e. a wrist camera. 
Using wrist camera images thereby facilitates cross-embodiment transfer, which may explain the predominancy of using them to provide inputs for VLA models~\cite{black2025pi05}.
We are not aware of ablation studies that measure the impact of attaching a camera to the wrist rather than elsewhere. But there is preliminary evidence that VLAs prioritize wrist cameras when picking up a tabletop object~\cite{wang2025pi0wild}.

As discussed in \Cref{sec:benchmarking}, there are several efforts to standardize robotic platforms~\cite{khazatsky2024droid}, aimed at facilitating transfer and benchmarking of VLA models between labs. 
Challenges related to cross-embodiment are thereby evaded rather than solved, highlighting how challenging it remains in practice.

\section{Collecting Action Data}
\label{sec:data}

Whereas the text and image data required to train LLMs and VLMs were already available in abundance on the Internet, action data required for visuomotor policy learning and VLAs remains scarce, and dedicated data gathering efforts are necessary~\cite{collaboration2024openx,khazatsky2024droid}.
This section provides examples of common data collection approaches. For a more complete overview, we refer to Section VI.A of~\cite{kawaharazuka2025vla}.

Early diffusion policy (DP) formulations
achieved strong performance with hundreds of demonstrations in single-task settings~\cite{chi2024diffusionpolicy}. 
As diffusion and flow-matching models are scaled to multi-task settings \cite{trilbmteam2025carefulexaminationlargebehavior} and integrated into VLA architectures \cite{black2025pi05}, data requirements increase substantially. Recent large-scale studies using diffusion transformer architectures report thousands of demonstrations, corresponding to nearly 2,000 hours of teleoperated data \cite{trilbmteam2025carefulexaminationlargebehavior}, while popular VLAs operate at data collection scales on the order of tens of thousands of hours \cite{black2024pi0}.

This highlights an important trend: DPs can be relatively data-efficient for narrowly defined tasks, but achieving broad generalization across tasks, embodiments, and environments requires significantly larger and more heterogeneous datasets.
Accordingly, different robot data collection methods exist, offering varying degrees of scalability and potential for transfer learning.

\begin{description}

\item[Teleoperation] involves a human operator controlling a physical robot to perform the target task, where all the sensory input and actions are recorded in the process.
This method often yields the most relevant data for the specific robot, but transfer to robots with different embodiments typically requires additional abstraction, retargeting, or shared representations.
Robot teleoperation systems commonly fall into two categories: \textit{Puppeteering}~\cite{zhao2023learning, wu2024gello,arunachalam2022holodex}, where operators directly control robots via specialized interfaces, offering intuitive demonstration of complex movements and direct perception of robot joint limits, but requiring specific hardware, training, and physical effort. 
\textit{Teleoperation in Virtual Reality (VR)} \cite{khazatsky2024droid,ding2024bunny} uses VR headsets combined with position-tracked handheld devices to control physical or virtual robots but the lack of haptic feedback often hinders fine control.

\item[Wearable Devices,] like handheld grippers~\cite{song2020grasping,chi2024universal, shafiullah2023bringing,chi2024universal}, motion capture gloves~\cite{zhang2025doglovedexterousmanipulationlowcost, wang2024dexcap}, and augmented reality glasses~\cite{kareer2024egomimic}, aim to capture natural human movements, and have
the potential to transfer to different robot embodiments. However,
mapping human motion and kinematics to the robot can be complex and is not
always possible due to workspace, speed and force limitations. Handheld grippers offer simple and direct control for
grasping and manipulation tasks, provide haptic feedback during the
demonstration, and are generally portable. However, their applicability is
limited to tasks that can be performed with such simple grippers, and certain gripper motions might be difficult to reproduce kinematically on the robot. Motion
capture gloves~\cite{wang2024dexcap} capture dexterous movements but face
retargeting challenges between human and robot hands. AR glasses record
egocentric views and hand movements, enabling natural interaction
capture, but pose difficulties in accurate tracking and retargeting,
often requiring significant teleoperation data for fine-tuning.
\end{description}

Instead of collecting new datasets, recent works leverage pre-trained VLMs to automatically relabel existing robot and human demonstration data. Instead of relying solely on manual annotations, these models provide rich semantic labels, affordances, or goal descriptors based on visual and contextual understanding \cite{zawalski2025ecot,belkhale2024rt,black2025pi05,geminiroboticsteam2025}. This relabeling allows researchers to reinterpret broad datasets under new tasks or goals, enabling more flexible, scalable learning pipelines and unlocking generalization across tasks. These may be combined with orthogonal approaches for reducing the number of demonstrations required to learn new tasks~\cite{vosylius2025instantpolicy}.

\section{Benchmarking Transfer in the Age of Foundation Models}
\label{sec:benchmarking}

Robotic systems that are built around the large pre-trained architectures discussed in
\Cref{sec:gradient_stopping} and \ref{sec:in_painting} are able to achieve strong zero-shot
performance on novel tasks, and improve further with
fine-tuning~\cite{singh2025ogvla,black2025pi05,amin2025pi06vlalearnsexperience,nvidia2025gr00t}.
However, this shift also introduces ambiguity. When models are pretrained on massive, weakly curated
datasets, it becomes challenging to define what counts as transfer, determine if the model has truly
generalized to a novel domain, and measure transfer.

Most existing robotics benchmarks were designed to measure task-specific performance rather than generalization, and typically fall into three paradigms:
\textit{Offline datasets} enable reproducible learning, for example on large amounts of grasping or pushing data, but offer limited insight into generalization beyond the training distribution and across domains~\cite{9560844,lum2024get, 7758091}.
\textit{Robotics challenges} such as the DARPA challenges or the Amazon Picking Challenge have accelerated progress in real world deployment. However, they typically involve fixed constraints and narrow task definitions limiting reproducibility and variation.
\textit{Closed-loop lab evaluations} provide controlled environments for testing physical systems, but object sets, initial conditions, and success criteria are often fixed and under-specified, limiting reproducibility and variation~\cite{kressgazit2024robotlearningempiricalscience,dasari2021rb}.
One notable recent effort toward principled evaluation is the work by the TRI team \cite{trilbmteam2025carefulexaminationlargebehavior}, discussed in Section \ref{sec:visuomotor},  which places a strong emphasis on performance in \textit{unseen} scenarios and on explicitly quantifying performance gains and losses resulting from fine-tuning. %

These paradigms are valuable for system validation but are not inherently designed to measure transfer. They rarely define clear source–target splits, test performance on truly novel domains, or provide metrics to evaluate what prior knowledge contributes to generalization -- whether from perception, policy structure, embodiment familiarity or task similarity. 

Thus, despite growing interest in VLAs, the field lacks understanding of when and how transfer works, as well as clear methodologies for evaluating it. In the following, we therefore offer recommendations for how the field can support this evolving direction.

\subsection{Towards Best Practices for Benchmarking Transfer}
There are general best practices for evaluating learned policies such as evaluators who are blind to the executed policy or a way to meticulously control initializations~\cite{kressgazit2024robotlearningempiricalscience,atreya2025roboarenadistributedrealworldevaluation}. To advance transfer learning in robotics, future benchmarks should additionally adopt the following principles:

\paragraph{Define explicit transfer splits and use transfer-oriented metrics.}
Benchmarks should specify source–target partitions across objects, tasks, and robots, and annotate novelty levels or at least measure similarity (e.g., unseen category, composition, morphology). For foundation models, pretraining data coverage should be documented to avoid inadvertent overlap.

With VLAs that are built around VLMs which in turn are trained across thousands of object and environment types, it is unclear whether it remains feasible to clearly label and partition object sets and environment sets and to evaluate generalization under controlled shifts (e.g., from household to industrial items). Works like $\star$-gen~\cite{taxonomy2025arxiv} set an example of how this can be done by defining a clear taxonomy of generalization, including the visual axis spanning object and environment changes. They train on an open-source dataset and define a methodology for evaluation through perturbations from base tasks.

Benchmarks should standardize metrics that reflect transfer learning goals -- including zero-shot success, few-shot curves, transfer gain, and retention. Evaluations should include both raw performance and adaptation cost.

\paragraph{Support modular and compositional evaluation.}
Systems often transfer at the level of subsystems -- e.g., perception, policy, or goal conditioning. Benchmarks should allow component-level evaluation and support compositional tasks testing skill reuse or novel goal composition. In VLA architectures that follow the layers in \Cref{fig:transfer_layers} -- especially those that use knowledge insulation -- the different layers can be ``unit tested'' first before performing integration tests at the overall system level.

\paragraph{Promote transparency and reusability.}
Community engagement -- through shared task definitions, open asset libraries, and reproducibility challenges -- will be critical in sustaining and scaling efforts to achieve transparency and reproducibility in benchmarking.
RoboArena~\cite{atreya2025roboarenadistributedrealworldevaluation} seems like a good first step in this direction by proposing a distributed methodology for comparing VLAs that relies on crowd-sourcing. The main requirement is pairwise execution of two policies on a specific robot platform, with results collected on a central server. While this still relies on evaluations in the real world that are riddled by challenges~\cite{zhang2025doglovedexterousmanipulationlowcost}, it alleviates the need to define tasks and domains. Combining this paradigm with simulation-based evaluation~\cite{geng2025roboverse, kawaharazuka2024review_new,trilbmteam2025carefulexaminationlargebehavior} could be a major step forward to scale up evaluations.
Furthermore, novel \textit{coopetitions}, as conducted in the euROBIN project~\cite{eurobin_project}, are competitions in which rewards are provided not only for task performance, but also for cooperation and achieving embodiment transfer between robotic teams.

\section{Conclusion}
\label{sec:conclusion}

Foundation models and transformer networks have constituted a step-change for transfer and generalization, and have brought robotics closer to mastering \textit{full-stack transfer} than ever before.
In considering foundation models from the perspective of transfer learning, several trends appear to indicate that the layers in \Cref{fig:transfer_layers} provide a good \textit{inductive bias}~\cite{toelle2025towards,marshall2025aretransformers} for organizing VLA architectures.
For instance, the layers \verbalinstructions, \goalspecifications, and \movementprimitives are clearly identifiable in current state-of-the-art VLAs such as \phin{0.5} and \phin{0.6}. And the importance of knowledge insulation highlights that care must be taken how information spreads from lower to higher layers during learning. More generally, decoupling the \textit{what} from the \textit{how} -- known as the \textit{two-stream hypothesis} in cognitive science~\cite{goodale1992twostreamhypothesis} -- appears to be an important strategy for making the most of sparse data~\cite{shridhar2022cliport}.

One current trend is the improvement of robotic skills in VLA architectures through reinforcement learning~\cite{amin2025pi06vlalearnsexperience}. The transfer layers are again apparent in~\cite{amin2025pi06vlalearnsexperience}, as the \goalspecifications and \movementprimitives are updated independently, but from the same value function.

Marshall and Barron~\cite{marshall2025aretransformers} argue that feedforward transformer-based applications are structurally incapable of reliable metacognition, and thus not able to achieve true autonomy. To improve the planning abilities and autonomy of robots, \textit{world models} have been proposed. In machine learning, the term ``world model'' refers to models that predict future world states from current states and actions~\cite{Ha2018WorldModels}.
These models can provide latent action spaces for VLAs~\cite{kawaharazuka2025vla}, and can leverage human and robot data to transfer knowledge to previously-unseen tasks~\cite{goswami2025worldmodelsleveragehuman,bharadhwaj2024gen2act}.
Real robot experiments report high zero-shot transfer in reaching, gripper grasping~\cite{assran2025vjepa2} and dexterous grasping tasks~\cite{goswami2025worldmodelsleveragehuman}, where a benchmark against one VLA hinted at higher generalization capabilities~\cite{assran2025vjepa2}. 
In robotics, the term ``world model'' instead refers to a centralized module that maintains the robot's belief state~\cite{triggs1991oxfordrobotworldmodel,sakagami23robotic}. These models are important for long-term task planning~\cite{pohl2024makeable} and efficiently storing and querying robot experiences~\cite{plewnia2024forgetting}; abilities current VLAs mostly lack.

How can reinforcement learning, world models, and other approaches and modules best be integrated into VLA-based architectures? 
These are some of the open questions that the community is addressing today. 
Our conclusion is that foundation models alone may not be \textit{the} route to full-stack transfer. But we certainly expect them to have and keep a primary role \textit{on} that route.
Distinguishable transfer layers and knowledge insulation are concepts that have recurrently emerged in different forms, as we have sought to highlight in this article. 
This convergent evolution suggests that that these concepts may well be effective design principles along this route.

\section*{Acknowledgments}
We sincerely thank the anonymous reviewers for feedback on an earlier version of this manuscript. 
F.S., S.B., J.S. and A.A.-S. were funded in part by the
European Union’s Horizon Research and Innovation Programme, grant 101070596 (euROBIN) and the Bavarian Ministry of Economic Affairs, Regional Development and Energy, through the project SMiLE-AI (VLR-2506-0002).
J.B. was funded in part by the Toyota Research Institute, NSF Award \#2327974, the Sloan Foundation, Stanford Human-Centered AI Institute, and Intrinsic.
S.S. was funded by NSF Awards \#2143601, \#2037101, and \#2132519.

\end{document}